\documentclass[letterpaper, 10 pt, conference]{ieeeconf}  

\usepackage{graphicx}
\usepackage{amsmath}
\usepackage{amsfonts}
\usepackage{algorithm}%
\usepackage{algorithmicx}%
\usepackage{algpseudocode}%
\usepackage{xcolor}%
\usepackage{booktabs}
\usepackage{hyperref}

\IEEEoverridecommandlockouts                              

\title{\LARGE \bf
VCAT: Vulnerability-aware and Curiosity-driven Adversarial Training for Enhancing Autonomous Vehicle Robustness
}

\author{Xuan Cai$^{1}$, Zhiyong Cui$^{*1}$, Xuesong Bai$^{1}$, Ruimin Ke$^{2}$, Zhenshu Ma$^{1}$, Haiyang Yu$^{1,3}$ and Yilong Ren$^{*1,3}$
\thanks{*This work was supported by the National Key Research and Development Project of China under Grant 2022YFB4300400 and the Beijing Natural Science Foundation (project number: L243008). (\textit{Corresponding author: Zhiyong Cui and Yilong Ren})}
\thanks{$^{1}$\{Xuan Cai, Zhiyong Cui, Xuesong Bai, Zhenshu Ma, Haiyang Yu and Yilong Ren\} is with State Key Laboratory of Intelligent Transportation Systems, School of Transportation Science and Technology, 
Beihang University, Beijing, 100191, P.R.China 
        {\tt\small (E-mail: \{caixuan, zhiyongc, xs\_bai, mzs0822, hyyu, yilongren\}@buaa.edu.cn)}}%
\thanks{$^{2}$Ruimin Ke is with Department of Civil and Environmental Engineering, Rensselaer Polytechnic Institute, Troy, New York, 12180, USA
        {\tt\small (E-mail:ker@rpi.edu)}%
}
\thanks{$^{3}$\{Haiyang Yu and Yilong Ren\} is with Zhongguancun Laboratory, Beijing, 100191, P.R.China
}}

\begin{document}

\maketitle
\thispagestyle{empty}
\pagestyle{empty}

\begin{abstract}

Autonomous vehicles (AVs) face significant threats to their safe operation in complex traffic environments. Adversarial training has emerged as an effective method of enabling AVs to preemptively fortify their robustness against malicious attacks. Train an attacker using an adversarial policy, allowing the AV to learn robust driving through interaction with this attacker. However, adversarial policies in existing methodologies often get stuck in a loop of overexploiting established vulnerabilities, resulting in poor improvement for AVs. To overcome the limitations, we introduce a pioneering framework termed Vulnerability-aware and Curiosity-driven Adversarial Training (VCAT). Specifically, during the traffic vehicle attacker training phase, a surrogate network is employed to fit the value function of the AV victim, providing dense information about the victim's inherent vulnerabilities. Subsequently, random network distillation is used to characterize the novelty of the environment, constructing an intrinsic reward to guide the attacker in exploring unexplored territories. In the victim defense training phase, the AV is trained in critical scenarios in which the pretrained attacker is positioned around the victim to generate attack behaviors. Experimental results revealed that the training methodology provided by VCAT significantly improved the robust control capabilities of learning-based AVs, outperforming both conventional training modalities and alternative reinforcement learning counterparts, with a marked reduction in crash rates. The code is available at \href{https://github.com/caixxuan/VCAT}{https://github.com/caixxuan/VCAT}.

\end{abstract}

\section{INTRODUCTION} 

AVs have gradually increased their market presence but have also become one of the sources of threats to public safety \cite{sun2021survey}. However, it is extremely challenging to comprehensively enhance the robustness due to sparse corner cases. Adversarial training provides an effective method \cite{im2022adversarial}. By allowing attackers, i.e., traffic vehicles, to create safety-critical scenarios, learning-based AVs are expected to learn how to avoid risks under safety expectations, thereby further enhancing robustness. In general, existing adversarial training methods face two challenges: insufficient utilization of the victim's intrinsic information and the limited variety of the attacker's attack modes. 

\subsection{Problems and Challenges}

\textbf{Exploitation of intrinsic vulnerability of victim}. Prevailing studies often utilize fused environmental observation via optimization \cite{han2020metamorphic} or learning \cite{tuncali2019requirements} methods to pinpoint the desired attack, while often neglecting the exploitation of the victim (i.e., target AV)'s intrinsic vulnerabilities. This oversight is consequential; reliance on mere observational data can yield substantial pitfalls, as attackers may struggle to identify unfavorable states of the black-box victim, making it difficult to launch effective attacks, particularly under conditions where safety-critical frames are rare. Such occurrences are quite common in AVs where the "long-tail effect" \cite{liu2024curse} exists. 

\textbf{Exploration of policy space of victim}. Traditional attack methods might only set binary collision or not, or a continuous probability distribution \cite{chen2024safety}. However, such tactics may falter due to inadequate exploration, leading to a phenomenon known as mode collapse, particularly under conditions of sparse rewards \cite{aradi2020survey}. This vulnerability is often exacerbated by the propensity for local optimization intrinsic to learning-based techniques.

\subsection{Main Contribution}

To address the above issues, we propose the VCAT framework, with its key contributions summarized as adversarial training framework, attack method, and rigorous experimentation. 

\begin{itemize}
\item {\textbf{Adversarial Training Framework}: VCAT. We have constructed a \textbf{v}ulnerability-aware and \textbf{c}uriosity-driven \textbf{a}dversarial \textbf{t}raining (VCAT) framework. This framework exploits identified weaknesses within the AV to fabricate a diverse spectrum of scenarios. Consequently, it enhances the AV’s competency in acquiring robust defensive driving strategies when faced with critical edge cases.}

\item {\textbf{Attack Method}: Inspired by the victim-aware and curiosity \cite{oudeyer2018computational} mechanism, we have developed a curiosity-driven deep reinforcement learning (DRL) attack paradigm, that leverages vulnerabilities of the victim by focusing on areas that the attacker has not fully understood or explored.}

\item {\textbf{Adversarial Training Experiment}: To rigorously evaluate the effectiveness of the VCAT framework, we conducted extensive adversarial training simulations. The results of these experiments reveal that our proposed method markedly bolsters the risk mitigation capabilities of AVs, thereby substantially elevating the safety standards in autonomous driving.}

\end{itemize}

\subsection{Construction}

The overall structure of the paper is as follows. Section II reviews related research work on DRL-powered attack and adversarial training. In Section III, we propose the VCAT framework, following a two-stage approach of adversarial attack and defense training \cite{xu2020adversarial}. Subsequently, in Section IV, the proposed method is conducted in a simulation experiment, and the results are analyzed. Finally, the conclusion and future works are summarized in Section V. Some commonly acronyms are also adopted, including \textit{w.r.t.} (with respect to) and \textit{w.l.o.g.} (without loss of generality). 

\section{RELATED WORKS} 

\subsection{DRL-powered Attack}
Attack methods employing DRL have accumulated substantial academic achievements by teaching adversarial agents to launch attacks. Especially in the field of AVs, artificial intelligence (AI) attacking AI is a common way. Through adversarial training, one can enhance the robustness of the target AI agent, a concept commonly seen in Generative Adversarial Network (GAN) \cite{aggarwal2021generative}, Generative Adversarial Imitation Learning \cite{ho2016generative}, and Game Theory \cite{owen2013game}.

In response to the limitations of traditional adversarial DRL, some literature aims to improve the performance in specific autonomous driving adversarial training and validation tasks. For instance, the series RL method proposed by Cai et al. \cite{cai2024adversarial} considerably diversified the range of adversarial scenarios. Huang et al. \cite{huang2022robust} leveraged Stackelberg game dynamics by factoring in the adaptivity of the agent, generating challenging yet solvable environments, thus enhancing the stability and robustness of RL training. 

Despite extensive research suggesting that constructing adversarial environments with RL aids in the training and validation, the benefits of integrating vulnerability-evaluation and curiosity-exploration of adversarial algorithms in learning tasks remain to be investigated. 

\subsection{Adversarial Training}
Adversarial training is a crucial method for enhancing the robustness of AI agents \cite{qayyum2020securing}, and it has accumulated substantial empirical research. The perspectives on adversarial phenomena can be dichotomized into adversarial attack and defensive strategies, or alternatively, they can be synthesized within an integrated framework of adversarial training. In terms of adversarial attacks, Ding et al. \cite{ding2021perceptual} devised a generative adversarial network aimed at stabilizing adversarial training to enhance contextual prediction in AVs through the restoration of visually degraded images. Kloukiniotis et al. \cite{kloukiniotis2022countering} reviewed denoising techniques as a countermeasure to adversarial attacks on AVs, emphasizing the role of adversarial training in improving adversarial robustness. In response to adversarial attack methods, adversarial defense is essential. Zhang et al. \cite{zhang2023cat} introduced a closed-loop adversarial training framework aimed at improving the robustness and safety of AV control. 

However, existing adversarial training methods have not exploited the intrinsic vulnerability nor explored the policy space of the victim, which hinders the advancements in the robustness of AI-driven AVs. 

\section{PROPOSED METHOD}

This section introduces the novel VCAT method, devised to bolster the safety of AVs via adversarial training. It first provides an overview of the VCAT framework and then elaborates on the proposed adversarial attack and defense protocols. 

\subsection{Overview of the VCAT Framework}

\begin{figure*}[ht!]
\centering
\includegraphics[width=18cm]{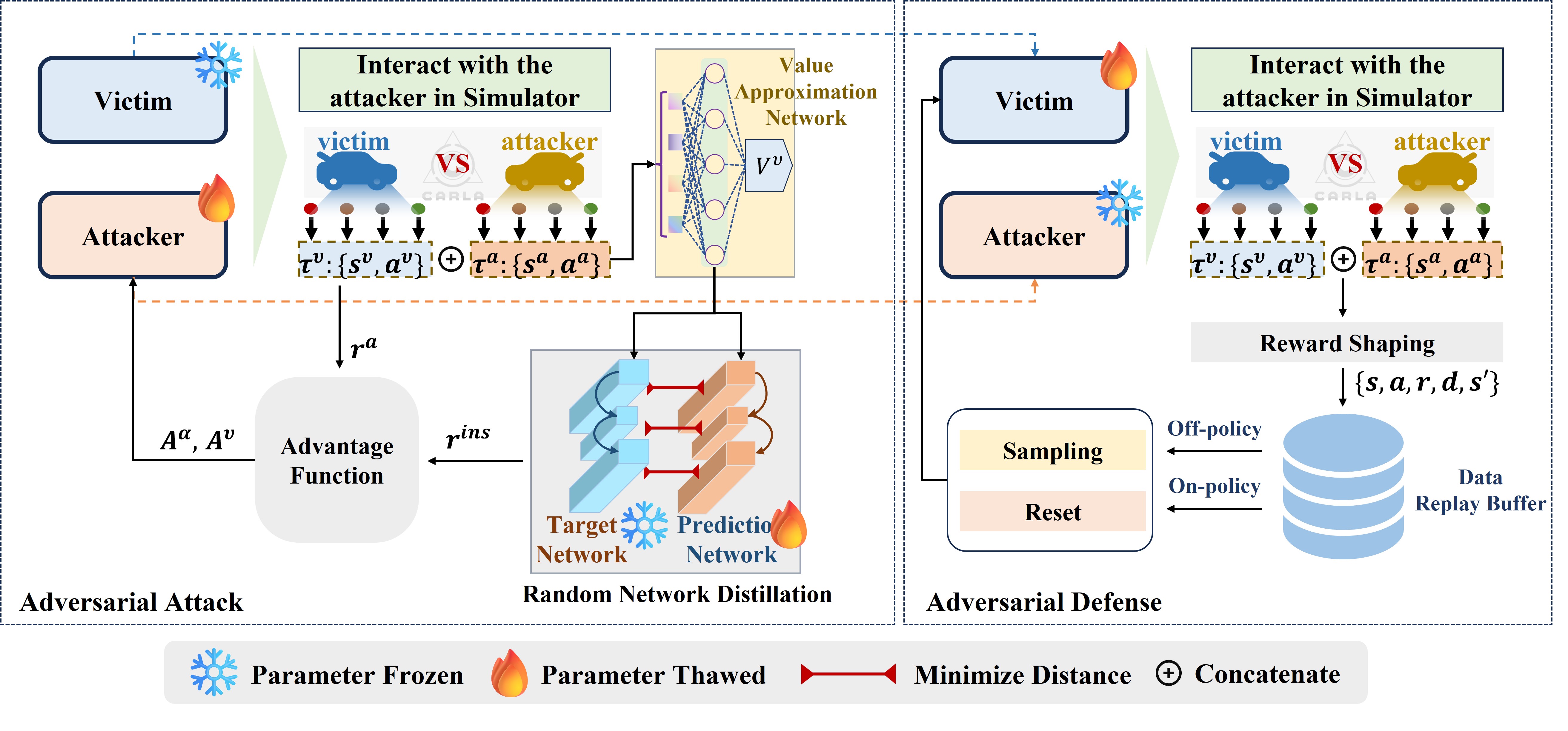}
\caption{Overview of the proposed VCAT framework. VCAT is divided into two stages: adversarial attack, enclosed by the left dashed box, and adversarial defense, enclosed by the right one. The snowflake pattern indicates that the neural network parameters are frozen, while the flame pattern indicates that the parameters can be adjusted for learning. The horizontal line with the reverse triangle arrow represents the minimum Euclidean distance between the two. The circled cross signifies data concatenation.}
\label{overview}
\vspace{-0.5cm}
\end{figure*}

The overview of the proposed VCAT framework is illustrated in Fig.\ref{overview}. It divides into dual stages of adversarial attack and defense, based on the victim's state, which alternates between being fixed (frozen) or variable (thawed) during the training and evaluation phases. In essence, the adversarial training studied in this paper models the game between the attacker and the victim as a two-player Markov Game (MG), which models the strategies of agents as part of the Markov Decision Process. In MGs, multiple agents perform a series of actions to maximize their collective or individual benefits. Specifically, two-player zero-sum MGs \cite{bansal2017emergent} involve a pair of agents with completely opposite interests. This study relaxes the zero-sum game problem due to the complicated traffic interactions. 

\subsection{Adversarial attack}

Before conducting adversarial attacks, it is imperative that the victim (target AV) be subject to extensive training using standard datasets (e.g., road-collected or random-generated data) to ascertain it possesses fundamental navigational proficiencies, albeit with a deficiency in managing anomalous or edge cases. Once the AV agent has been thoroughly trained, its parameters should be frozen to play the role of the victim, $\upsilon$, thus being attacked by the training attacker, $\alpha$. 

Victim and attacker constitute a two-player MG. When both are RL-driven, their value functions are $V_{\pi_{\theta_\alpha}}^\upsilon(s)$ and $V_{\pi_{\theta_\alpha}}^\alpha\left(s\right)$, also known as expected rewards \cite{hao2023adversarial}. Therefore, the goal of adversarial attack is for the attacker to learn to adeptly discern and exploit the victim's vulnerabilities, specifically by minimizing $V_{\pi_{\theta_\alpha}}^\upsilon(s)$. Given the network parameters $\theta_\upsilon$ remain frozen, policy $\pi_{\theta_\upsilon}$ is thereby fixed, which effectively incorporates the victim as an integral component of the environmental construct. Thus, the objective of the adversarial attack is quantified as: 
\begin{equation}
\label{Eq1}
    J=arg \max_{\theta_{\alpha}}\left(V_{\pi_{\theta_{\alpha}}}^{\alpha}(s)-V_{\pi_{\theta_{\alpha}}}^{v}(s)\right)
\end{equation}

Therefore, an important insight is that if we can estimate the victim's $V_{\pi_{\theta_\alpha}}^\upsilon(s)$, it would help find its weaknesses more accurately. The Proximal Policy Optimization (PPO) paradigm \cite{hao2023adversarial} is used to train the $\pi_{\theta_\alpha}$.

\subsubsection{Victim Value Approximation Network}

We use an approximation network (parameterized by $\theta_\upsilon$) to fit the state-value function of $\upsilon$, which aids in the explicit formulation of Eq.\ref{Eq1}. Adopting the Temporal Difference (TD) learning paradigm, we define the loss function of the approximation network equivalent to the TD-error: 
\begin{equation}
    arg\min_{\theta_\upsilon}\left\|V_{\pi_{\theta_\alpha}}^\upsilon(s_t)-\left(\widehat{r^\upsilon}(s_t,a_t)+\gamma\mathbb{E}_{s_{t+1}\sim P}[V_{\pi_{\theta_\alpha}}^\upsilon(s_{t+1})]\right)\right\|^2
\label{Eq3}
\end{equation}
where $\gamma$ is the discount factor, $P$ is the state transition probability, $a_t=(a_t^\alpha,a_t^\upsilon)$ is the sampled joint action, and $\widehat{r^\upsilon}$ is the estimated reward function for the victim, \textit{w.l.o.g.}, under the black-box assumption. This is done to extend the victim's generality, which is beneficial for comprehensive training and validation endeavors:
\begin{equation}
    \widehat{r^{\upsilon}}=\lambda_1\cdot r_{target}-\lambda_2\cdot r_{acc}-\lambda_3\cdot r_{collision}
\label{Eq4}
\end{equation}
where $\lambda$ is the weight, $r_{target}$ is the reward for the victim reaching the goal, $r_{acc}$ is the acceleration reward, and $r_{collision}$ is the collision reward. 

\subsubsection{Curiosity-Driven Exploration}

The value approximation network (VAN) is capable of approximating the significance of the current state $s_t$ \textit{w.r.t.} the victim; hence, $\theta_\upsilon$ encapsulates dense information about the state value of $\upsilon$ at the said $s_t$. If a certain $s_t$ represents an unfamiliar state to $\upsilon$, it becomes imperative to induce the adversarial agent to probe and exploit this state. Inspired by the random network distillation mechanism (RND) \cite{oudeyer2018computational}, two networks are constructed: a stationary target network, $\varrho$ and a dynamic predictor network, $\hat{\varrho}$. The parameter of $\varrho$ is randomized and then fixed, while $\hat{\varrho}$ is continuously optimized after randomization, with the aim of continuously approximating $\varrho$. This iterative optimization process is driven by the intent to minimize prediction disparities. When $\hat{\varrho}$ encounters a fresh state, the prediction error will be high, resulting in a high intrinsic reward output. Considering that the last hidden layer of $\theta_\upsilon$ (denoted as $\varphi^\upsilon)$ due to its potent representation of the characteristics of the dense state \textit{ w.r.t.} $V_{\pi_{\theta_\alpha}}^\upsilon$, it is used as input for RND. Therefore, the mean square error of RND is:
\begin{equation}
    r^{ins}=\|\hat{\varrho}(\varphi^{v}(s_{t}))-\varrho(\varphi^{v}(s_{t}))\|^{2}
\label{Eq5}
\end{equation}
where $r^{ins}$ is the intrinsic reward, which can adaptively adjust the exploration value of $s_t$ to steer the exploration of the attacker. 

\subsubsection{Attacker Policy Training}

The crux of the PPO lies in the calculation of the advantage function. \textbf{Algorithm \ref{algo1}} incorporates $r^{ins}$ into the advantage function, simultaneously coordinating $r^\upsilon$ and $r^\alpha$. This training requires the initialization of six networks. The calculations of the advantage functions $A_t^\alpha$, $A_t^\upsilon$, and $A_t^{\alpha,ins}$ are shown from lines 7 to 9. Subsequently, the training objective for PPO as in Eq.\ref{Eq3} can be computed: 
\begin{equation}
    \begin{aligned}
    \begin{cases}arg\max_{\theta_\alpha}\mathbb{E}_{(a_t^\alpha,s_t)\sim\pi_{\theta_{\alpha,k}}}[min\big(\rho_tA_t^\alpha,clip\big(\rho_t,1-\epsilon,1+\epsilon\big)\\\qquad\qquad\qquad A_t^\alpha\big)-min\big(\rho_tA_t^v,clip\big(\rho_t,1-\epsilon,1+\epsilon\big)A_t^v\big)\big]\\\rho_t=\frac{\pi_\alpha(a_t^\alpha|s_t)}{\pi_{\alpha,k}(a_t^\alpha|s_t)}\\
    \begin{cases}A_t^\upsilon=A_{\pi_{\alpha,k}}^\upsilon(a_t^\alpha,s_t)\\A_t^\alpha=A_{\pi_{\alpha,k}}^\alpha(a_t^\alpha,s_t)+\lambda A_{\pi_{\alpha,k}}^{\alpha,ins}(a_t^\alpha,s_t)&\end{cases}\end{cases}
\end{aligned}
\label{Eq6}
\end{equation}
where $\lambda$ denotes a hyperparameter that signifies the degree of exploration. This objective function is designed to leverage both the victim's value function and the intrinsic value of exploration, aiming to expeditiously navigate towards a state that maximizes expected rewards. Consequently, $r^{ins}$ can be internalized as the advantage function of PPO, with $\lambda$ balancing exploitation and exploration, and its value setting is referenced to \cite{gong2022curiosity}.

\begin{algorithm}[!bt]
\caption{Adversarial Attack}
\label{algo1}
\begin{algorithmic}[1]
\Require  $V_{\pi_{\theta_\alpha}}^\upsilon$: state value of the victim; $\varrho$: target network of the RND; $\hat{\varrho}$: predictor network of the RND; $V_{\pi_{\theta_\alpha}}^{\alpha,ins}$: state value of the intrinsic reward; $V_{\pi_{\theta_\alpha}}^\alpha$: state value of the attacker; $\pi_{\theta_\alpha}$: attacker policy; 
\For{$n=1,2,...,N$}
    \While {not done}
        \State $s_t=env.step(\upsilon,\alpha)$
        \State Collect trajectory: $\mathcal{T}.\boldsymbol{append}(s_t)$
    \EndWhile
    \State Compute $r_t^{ins}$ in each step of $\mathcal{T}$ \Comment{\textcolor{blue}{Based on Eq.\ref{Eq5}}}
    \For{$i=1,2,...,T$ in $\mathcal{T}$}
        \State $A_t^\alpha=r_t^\alpha+\gamma V_{\pi_{\theta_\alpha}}^{\alpha(t)}\left(s_{t+1}^\alpha\right)-V_{\pi_{\theta_\alpha}}^{\alpha(t)}\left(s_t^\alpha\right)$
        \State $A_t^\upsilon=r_t^\upsilon+\gamma V_{\pi_{\theta_\alpha}}^{\upsilon(t)}\left(s_{t+1}^\alpha\right)-V_{\pi_{\theta_\alpha}}^{\upsilon(t)}\left(s_t^\alpha\right)$
        \State $A_t^{\alpha,ins}=r_t^\upsilon+\gamma V_{\pi_{\theta_\alpha}}^{\alpha(t),ins}\left(s_{t+1}^\alpha\right)-V_{\pi_{\theta_\alpha}}^{\alpha(t),ins}\left(s_t^\alpha\right)$
    \EndFor
    \State Update $\pi_{\theta_\alpha}$ by minimizing the loss
    \Comment{\textcolor{blue}{Based on Eq.\ref{Eq6}}}
    \State Update $V_{\pi_{\theta_\alpha}}^\upsilon$, $V_{\pi_{\theta_\alpha}}^{\alpha,ins}$, $ V_{\pi_{\theta_\alpha}}^\alpha$ by minimizing the TD error
    \State Update $\hat{\varrho}$ by minimizing the loss \Comment{\textcolor{blue}{Based on Eq.\ref{Eq5}}}
\EndFor
\State \textbf{return} $\mathcal{T}$ 
\end{algorithmic}
\end{algorithm}

\subsection{Adversarial Defense}

Upon successful execution of the adversarial attack training, that is, once the policy governing the attacker has satisfied the pre-established criteria for convergence, the network parameters attributed to the attacker are henceforth frozen. Concurrently, the victim's parameters are thawed to learn defensive strategy against the onslaught of the well-trained attacker. Similarly, inverting Eq.\ref{Eq1} as follows:
\begin{equation}
    J=arg \min_{\theta_{v}}\left(V_{\pi_{\theta_{v}}}^{\alpha}(s)-V_{\pi_{\theta_{v}}}^{\upsilon}(s)\right)
\label{Eq7}
\end{equation}
where the parameters $\theta_\alpha$ are frozen, meaning $\pi_{\theta_\alpha}$ is fixed, while $\theta_\upsilon$ is thawed to learn to minimize Eq.\ref{Eq7}.

Note that the victim can be any construct, but the PPO is adopted as the model to assess the potency of the adversarial training.

\section{EXPERIMENT}

This study selects three scenarios for experiments. The simulation is conducted on a desktop PC equipped with a CPU Core i7 and a GPU NVIDIA 4070 Ti, using the highway-env \cite{highway-env}. This section details the experiment setup, research questions, results, and analysis.

\subsection{Experiment Setup}
\subsubsection{Scenario Setup}

\begin{figure}[bt!]
\centering
\includegraphics[width=8.5cm]{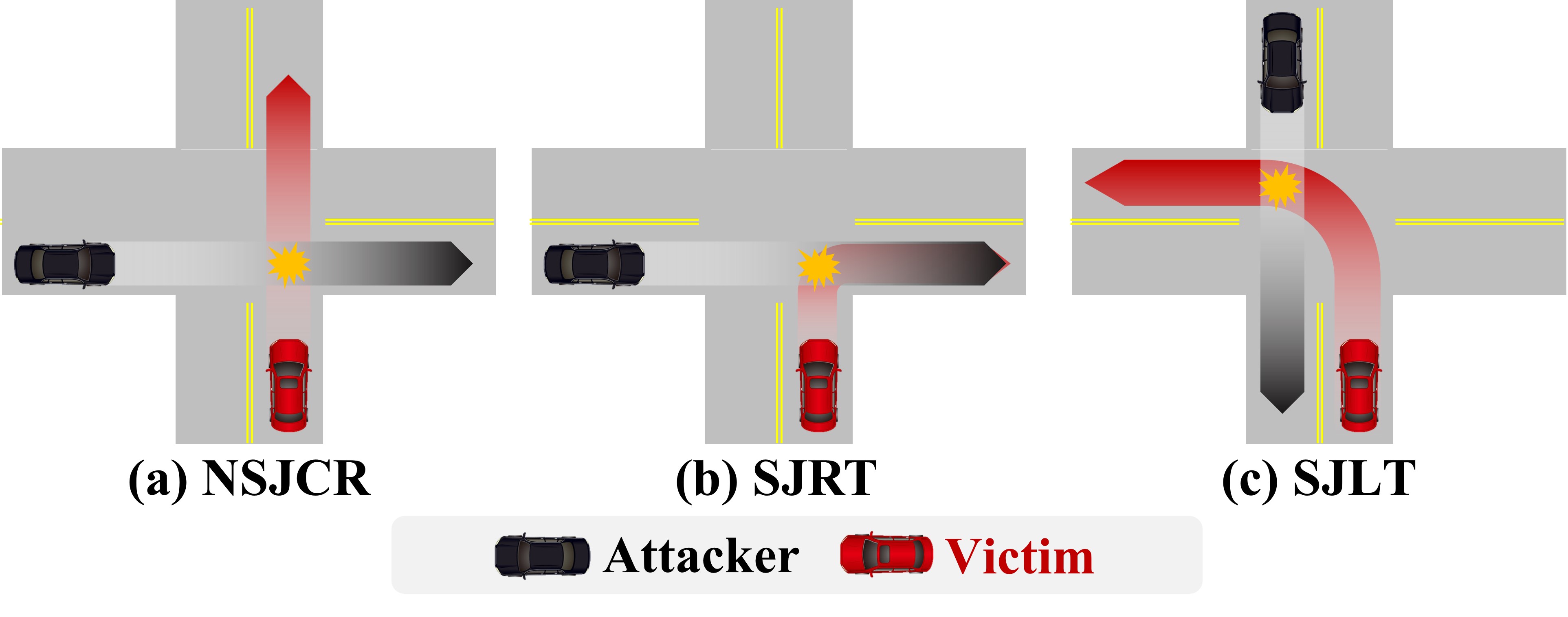}
\caption{Illustration for the setup of the three scenarios. The trajectory of the AV (victim) is represented by the red line, while the trajectory of the traffic vehicle (attacker) is represented by the black line. The scenarios are (a) \# NoSignalJunctionCrossingRoute (\# NSJCR), (b) SignalizedJunctionRightTurn (\# SJRT), and (c) SignalizedJunctionLeftTurn (\# SJLT), respectively. The abbreviations are used hereafter.}
\label{scenario_setup}
\vspace{-0.cm}
\end{figure}

The experiments set up three typical interactive scenarios, as illustrated in Fig.\ref{scenario_setup}, all of which are interactive dual-vehicle intersections that are recognized as hotspots for vehicular collisions. The black attacker (referred to as the traffic vehicle) is equipped with an adversarial protocol, $\pi_{\theta_\alpha}$, enabling it to methodically engineer safety-critical situations that challenge the response robustness of the victim (referred to as the target AV dominated by $\pi_{\theta_\upsilon}$).

\subsubsection{Hyper-parameter Setup}

The generalized training regimen of the victim before the adversarial attack is outside the scope of this study, and the key detail of the hyperparameters in this study is shown in Tab.\ref{tab:ParameterSetup1} referred to \cite{gong2022curiosity}\cite{tian2024enhancing}.

\begin{table}[bt!]
\centering
\caption{Hyper-parameter setup.}
\label{tab:ParameterSetup1}
\begin{tabular}{*{2}{p{3.9cm}}}
\toprule

\multicolumn{2}{c}{\textbf{PPO Attacker}} \\ \hline
buffer capacity & 5000 \\
batch size & 128 \\
learning rate of policy & 5.0e-4 \\
learning rate of value & 5.0e-3 \\
$\epsilon$ & 0.9 \\
train iteration & 10 \\
network dimension of policy & [state dim, 128, 64, action dim] \\
network dimension of value & [state dim, 128, 64, 1] \\
$\lambda$ (curiosity exploration) & 0.2 \\
$\gamma$ & 0.95 \\ \hline

\multicolumn{2}{c}{\textbf{Value Approximation Network}} \\ \hline
learning rate &	1.0e-3 \\
network dimension &	[input dim, 64, output dim] \\ \hline

\multicolumn{2}{c}{\textbf{Random Network Distillation}} \\ \hline
learning rate &	1.0e-3 \\
network dimension &	[input dim, 128$\times$3, output dim] \\

\bottomrule
\end{tabular}
\vspace{-0.cm}
\end{table}




\subsubsection{Baseline Setup of Adversarial Attack}

This paper selects several state-of-the-art methods as baselines, particularly focusing on the RL-based family that shares the same origin as the proposed method. For fair comparison, the reward or loss function is set to the same sparse modality. 

\begin{itemize}
\item{\textbf{Monte Carlo Sampling/Random (MC)} \cite{o2018scalable}: The initial state of the attacker within a limited area is set randomly.}
\item{\textbf{REINFORCE/Learning-to-Collide (LC)} \cite{ding2020learning}: The concept of GAN is utilized to generate safety-critical data.}
\item{\textbf{NormalizingFlow (NF)} \cite{ding2021multimodal}: The normalizing flow generator is leveraged to create natural and adversarial safety-critical data.}
\item{\textbf{RL-PPO} \cite{xu2022safebench} / \textbf{RL-DDPG} \cite{chen2021adversarial} / \textbf{RL-TD3} \cite{cai2024adversarial} / \textbf{RL-SAC} \cite{xu2022safebench}: RL-based agents are employed to play the role of attacker.}
\end{itemize}

\subsection{Research Questions}

Prior to the initiation of experimental procedures, we have articulated three research inquiries to steer the experimental design and execution:

\begin{itemize}
\item{\textbf{RQ.1.} What is the efficacy of the VCAT in supporting adversarial attacks?}
\item{\textbf{RQ.2.} Does the VCAT provide a superior level of resilience against adversarial maneuvers compared to others?}
\item{\textbf{RQ.3.} How does each component of the VCAT contribute to the attack capability (i.e., ablation studies)?}
\end{itemize}

\subsection{Experiment Result}
\subsubsection{RQ1. Efficacy of Adversarial Attack}

\textbf{Metrics}. The crash rate characterizes the efficiency of generating safety-critical collisions \textit{w.r.t.} the attack method \cite{o2018scalable}. A more rapid increase in the crash rate signifies greater efficiency. To measure the coverage of attack methods, t-SNE \cite{gong2022curiosity} is used to visualize all action vectors from the slice trajectories of the victim interacting with different attackers in 2-D space. The wider the coverage of t-SNE, the richer the behaviors activated by $\pi_{\theta_\upsilon}$, and the more vulnerabilities exposed. The number of crashes is another metric specifically used to measure the diversity of different types of edge scenarios \cite{feng2023dense}; the richer, the better. For the features of all the crashes, we distinguish four categories to examine the richness of the scenarios generated. 

\begin{figure}[bt!]
\centering
\includegraphics[width=8.5cm]{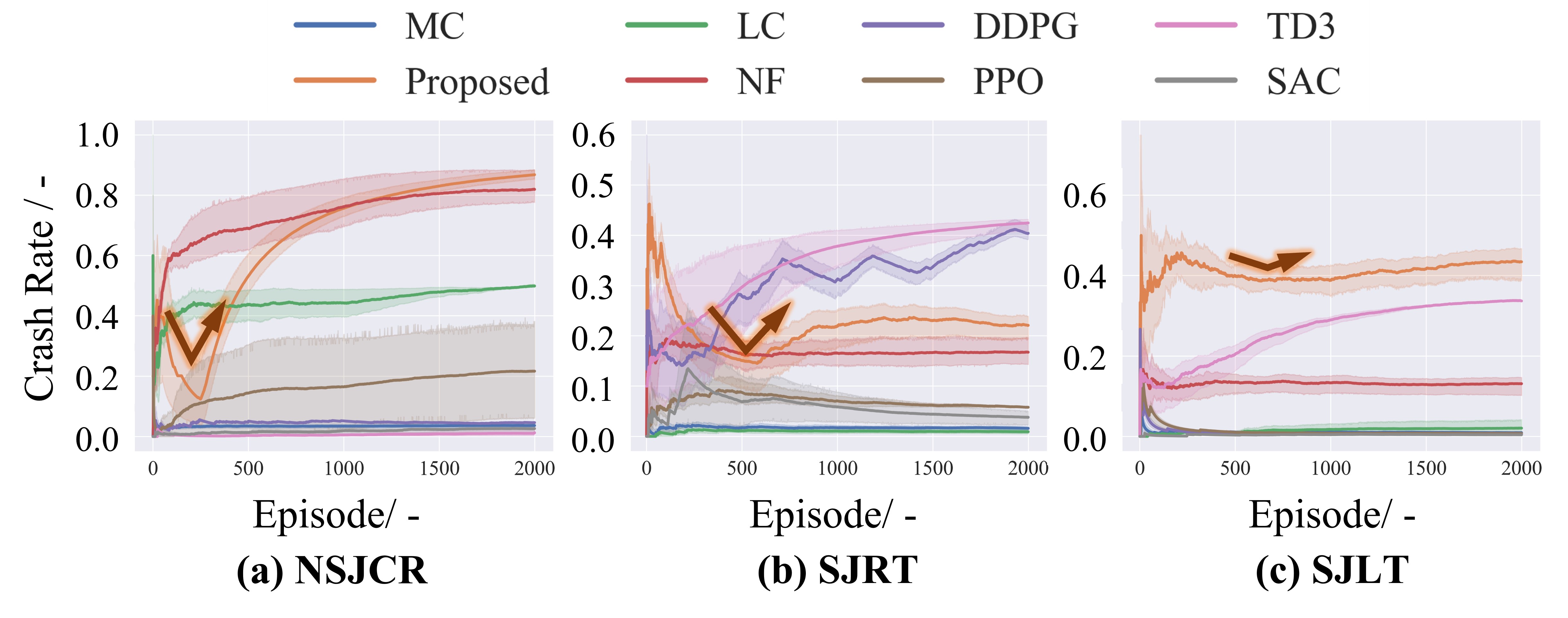}
\caption{Crash rate in the adversarial attack training with different methods. The orange "V"-shaped arrows highlight the decline-rise process experienced by the proposed method.}
\label{crash rate}
\vspace{-0.cm}
\end{figure}

\begin{figure}[bt!]
\centering
\includegraphics[width=8.5cm]{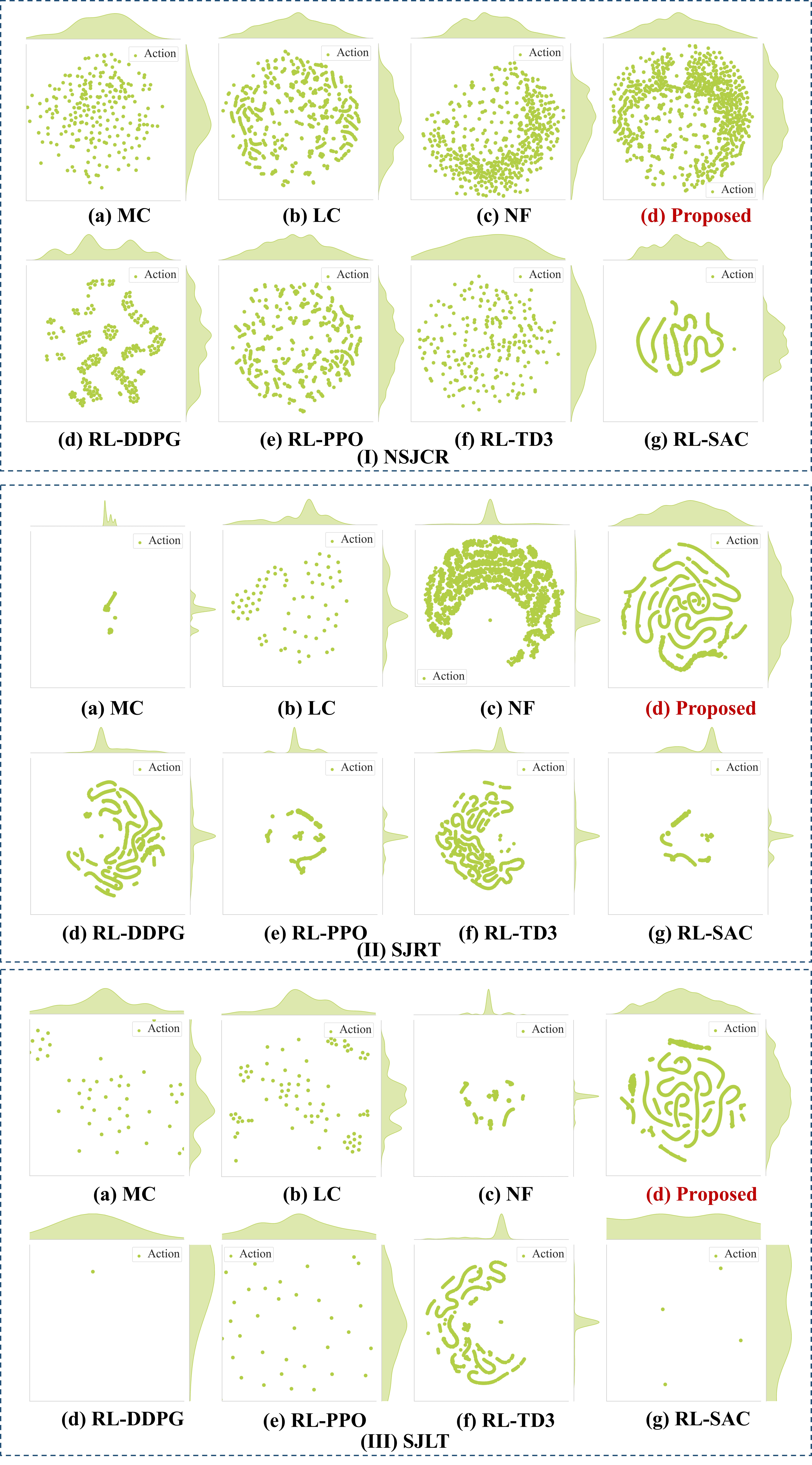}
\caption{t-SNE visualization of the victim (target AV) in the attack training under the three scenarios. The size of the coordinate axis is consistent for each scenario.}
\label{tSNE}
\vspace{-0.1cm}
\end{figure}

\begin{figure}[bt!]
\centering
\includegraphics[width=8.5cm]{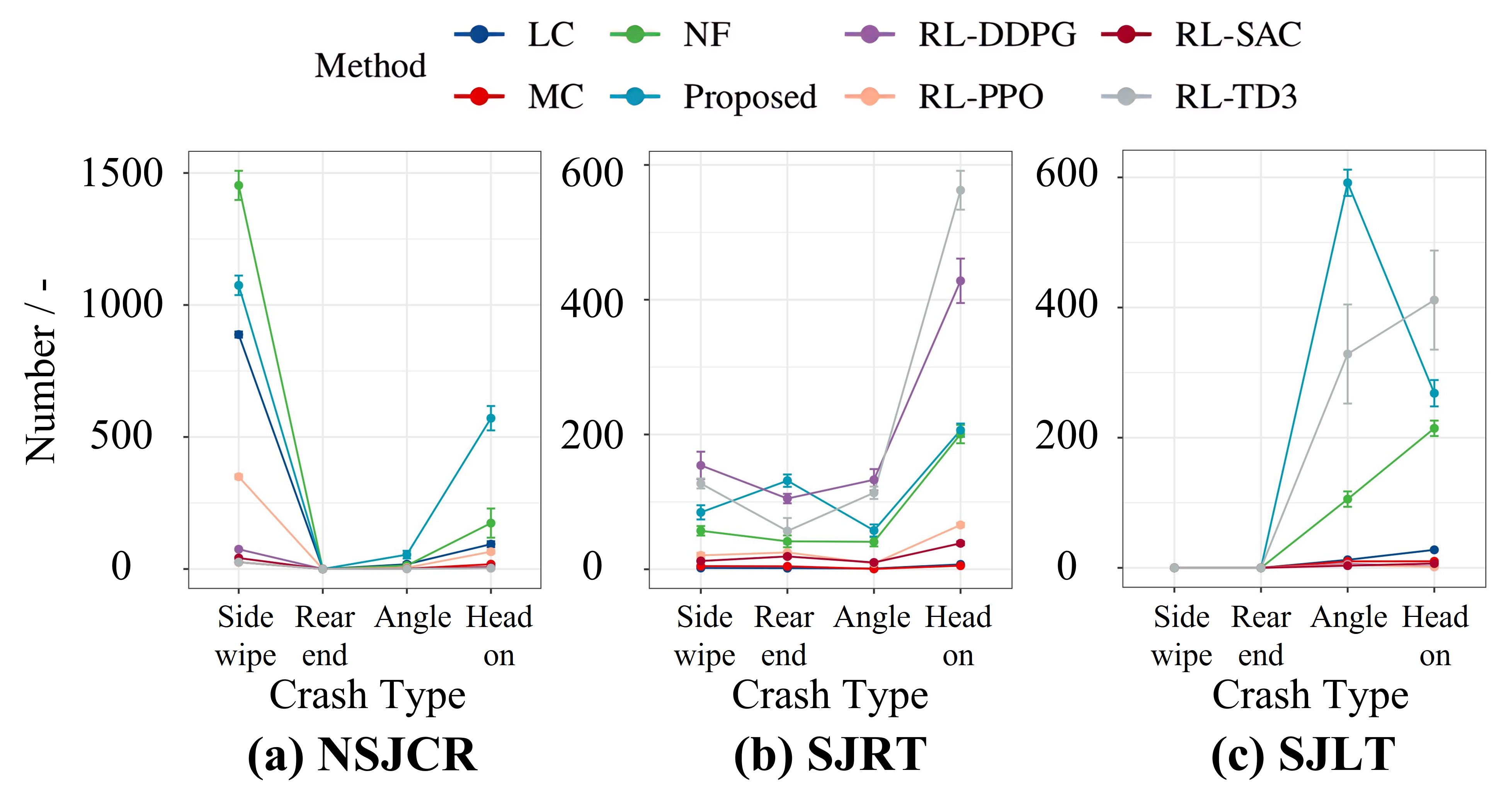}
\caption{Number of crashes ($\uparrow$) in different types under the three scenarios.}
\label{crash bar}
\vspace{-0.cm}
\end{figure}

\textbf{Results}. Fig.\ref{crash rate} shows the crash rates under the three scenarios. The following characteristics can be identified: 1) Many baseline methods struggle to form effective attacks with the sparse incentives, prone to mode collapse in the limited time, such as DDPG, PPO, SAC, etc., in the first scenario. The proposed method, however, can avoid this issue, with the crash rate rising to a high level. 2) The proposed adversarial attack method experiences a distinct "V"-shaped phase of decline followed by an increase during early stages, as emphasized by the orange V-shaped arrows. Fig.\ref{tSNE} presents the 2-D t-SNE visualization of the victim's action vector. It can be observed that the data distribution of the proposed is more widespread, suggesting that, compared to other counterparts, it can activate a richer policy within the victim, helping to uncover more vulnerabilities. Fig.\ref{crash bar} illustrates the number of crashes during adversarial attack training. The proposed method, although not the most prevalent in each category, exhibits the best average performance across the three scenarios. 

\textbf{Analysis}. The method introduced herein adeptly circumvents mode collapse and assimilates potent adversarial patterns, achieving a higher crash rate. The V-shaped feature in Fig.\ref{crash rate} and the extensive data distribution in Fig.\ref{tSNE} further demonstrate the enhanced exploration capability of our approach without the exploitation of the internal knowledge within the victim. Although the proposed does not consistently achieve the highest crash rate, as seen in the second scenario where it performs slightly worse than TD3 and DDPG, it improves the learning efficiency of RL under the sparse incentive condition, maintaining a balanced exploration and exploitation, especially suitable for such rare safety-critical conditions. For instance, DDPG exhibits mode collapse in the other two scenarios.

\subsubsection{RQ2. Comparison of Adversarial Training}

\textbf{Metrics}. Non-Crash Rate (as shown in Tab.\ref{tab:Non-crash rate}). Comparing the non-crash rate validated by different attack methods under various adversarial training methods, a higher non-crash rate is preferable. To test the effectiveness of adversarial training, cross-training and validation are employed. Taking the second column in Tab.\ref{tab:Non-crash rate} as an example, the adversarial training method uses MC, followed by the validation method using the proposed, to test whether the victim can withstand the attack from the proposed after being trained in MC. 

\textbf{Results}. We selected MC, DDPG, and TD3 as baselines and compared four cross-adversarial training and validation categories. 1) AT: MC+Val.: Prop.: Despite training under MC, the victim exhibits a lower non-crash rate when confronted with the proposed attack, suggesting inadequate training; 2) AT: DDPG+Val.: Prop.: Following adversarial training with DDPG, the non-crash rate generally increases compared to the MC one, except in the first scenario where it fails; 3) AT: Prop.+Val.: MC: Training with the proposed method results in a consistently high non-crash rate, indicating that the victim agent can effectively handle universal scenarios; 4) AT: Prop.+Val.: TD3: When the validation method is switched to TD3, the non-crash rate remains high, demonstrating that the proposed training method is robust against maliciously trained attacks. 

\textbf{Analysis}. The proposed method effectively uncovers a comprehensive attack space, encompassing a broader range of edge scenarios. Adversarial training with this approach significantly enhances the victim’s robustness, enabling it to effectively handle universal MC scenarios and resist TD3’s malicious attacks to a large extent. However, despite successful adversarial training, other methods exhibit limited policy action activation exploration, thereby constraining their generalization performance.

\begin{table}[bt!]
\centering
\caption{Non-Crash rate under different validation methods after adversarial training. (AT: Adversarial training; Val.: Attack methods used to validate adversarial training).}
\label{tab:Non-crash rate}
\begin{tabular}{*{5}{p{1.3cm}}}
\toprule
Non-Crash Rate ($\uparrow$) /\%	&AT:MC+ Val.:Prop.	&AT:DDPG+ Val.:Prop.	&AT:Prop.+ Val.:MC	&AT:Prop.+ Val.:TD3 \\ \hline
\#NSJCR	&16.4±2.6\%	&8.2±1.6\%	&98.0±1.1\%	&97.0±1.4\% \\
\#SJRT	&76.8±3.9\%	&82.0±7.0\%	&99.1±0.2\%	&89.2±6.7\% \\
\#SJLT	&57.1±5.4\%	&71.9±3.3\%	&97.3±1.7\%	&93.7±3.9\% \\
\bottomrule
\end{tabular}
\vspace{-0.2cm}
\end{table}

\subsubsection{RQ3. Ablation Studies}

We focus on the ablation studies of attacking efficacy. \textbf{Ablation baseline:}
\begin{itemize}
    \item {PPO: The raw PPO adversarial attack method;}
    \item {PPO-VA: Vulnerability-aware PPO, in which the curiosity exploration hyperparameter is set to zero, i.e., $\lambda=0$;}
    \item {Proposed: The full method introduced in this paper, $\lambda=0.2$.}
\end{itemize}

\textbf{Results}. The ablation experiment results are shown in Tab.\ref{tab:ablation}. MC is clearly inferior to the PPO method. However, the PPO still exhibits low attack efficiency, with a maximum of only about 21.6\%. When the vulnerability-aware module is incorporated, the improvement in the crash rate is minimal and even decreases, with a maximum increase of only about 1.3\%. When the proposed method is fully implemented, the crash rate significantly increases, particularly achieving a high crash rate of 83.4\% in the first scenario. 

\textbf{Analysis}. The utility of using the vulnerability-aware module alone is limited. This is because without the introduction of the exploration mechanism, it merely weights the states where the attacked victim may have vulnerabilities. However, some error exists in the estimated reward (see Eq.\ref{Eq4}), making it difficult to achieve improvements using only the VAN. The curiosity mechanism must be combined to explore a larger space; otherwise it will result in excessive exploitation.

\begin{table}[bt!]
\centering
\caption{Ablation studies for adversarial attack.}
\label{tab:ablation}
\begin{tabular}{*{5}{p{1.3cm}}}
\toprule
Crash Rate ($\uparrow$) /\%	&MC	&PPO	&PPO-VA	&Proposed \\ \hline
\#NSJCR	&1.2±0.2\%	&21.6±2.9\%	&22.2±2.4\%	&83.4±6.4\% \\
\#SJRT	&1.0±0.1\%	&7.3±1.4\%	&7.2±1.7\%	&23.8±3.2\% \\
\#SJLT	&1.9±0.4\%	&2.2±0.3\%	&3.5±0.3\%	&46.0±4.7\% \\
\bottomrule
\end{tabular}
\vspace{-0.1cm}
\end{table}

\section{CONCLUSIONS And FUTURE WORKS}

This paper proposes a vulnerability-aware and curiosity-driven adversarial training (VCAT) framework to overcome the challenge of adequately enhancing exploration while achieving a balance with exploitation, especially in the sparse, safety-critical scenarios. A pioneering adversarial training framework is constructed, consisting of two stages: adversarial attack and adversarial defense, to enhance the robustness of autonomous driving. In the adversarial attack phase, a vulnerability-aware and curiosity-driven module that enhances attack robustness and efficacy is introduced, enabling the traffic attacker to learn to generate sufficient rare safety-critical data. In the adversarial defense phase, the autonomous vehicle victim gradually learns how to defend against malicious attacks from the pretrained attacker through interactions. Experimental results demonstrated that the proposed adversarial training method can significantly better enhance the robustness of autonomous driving compared to other counterparts.

Future work will focus on incorporating real-world data into the training process, expanding the range of adversarial scenarios, and strengthening the system's resilience against adaptive adversaries.

\addtolength{\textheight}{-7.0cm}   








\bibliographystyle{unsrt}
\bibliography{ref}


\end{document}